\documentclass[10pt,twocolumn,letterpaper]{article}
\pdfoutput=1
\usepackage{iccv}
\usepackage{times}
\usepackage{epsfig}
\usepackage{graphicx}
\usepackage{amsmath}
\usepackage{amssymb}

\usepackage{subfigure}
\usepackage{multirow}
\usepackage{array}

\usepackage{booktabs}

\usepackage[breaklinks=true,bookmarks=false]{hyperref}

\iccvfinalcopy 

\def\iccvPaperID{2586} 

\ificcvfinal\fi

\begin{document}

\title{Personalized Trajectory Prediction via Distribution Discrimination}

\author{%
Guangyi Chen$^{1,2,3}$, Junlong Li$^{1,2,3}$, Nuoxing Zhou$^{1,2,3}$, Liangliang Ren$^{1,2,3}$, Jiwen Lu$^{1,2,3,}\thanks{Corresponding author}$\ \\
{$^1$Department of Automation, Tsinghua University, China}\\
{$^2$State Key Lab of Intelligent Technologies and Systems, China}\\
{$^3$Beijing National Research Center for Information Science and Technology, China}\\
{\tt\small chen-gy16@mails.tsinghua.edu.cn; \{ljlong.leo,nuoxingzhou\}@gmail.com;}\\
{\tt\small renliangliang@cvte.com; lujiwen@tsinghua.edu.cn}\\
}



\maketitle
\ificcvfinal\thispagestyle{empty}\fi

\begin{abstract}
   Trajectory prediction is confronted with the dilemma to capture the multi-modal nature of future dynamics with both diversity and accuracy. In this paper, we present a distribution discrimination (DisDis) method to predict personalized motion patterns by distinguishing the potential distributions. Motivated by that the motion pattern of each person is personalized due to his/her habit, our DisDis learns the latent distribution to represent different motion patterns and optimize it by the contrastive discrimination. This distribution discrimination encourages latent distributions to be more discriminative. 
   Our method can be integrated with existing multi-modal stochastic predictive models as a plug-and-play module to learn the more discriminative latent distribution. 
   To evaluate the latent distribution, we further propose a new metric, probability cumulative minimum distance (PCMD) curve, which cumulatively calculates the minimum distance on the sorted probabilities.
   Experimental results on the ETH and UCY datasets show the effectiveness of our method. $\footnote{Code and a video demo are available at \url{https://github.com/CHENGY12/DisDis}}$ 

\end{abstract}

\section{Introduction}
Human trajectory prediction aims at forecasting future pedestrian trajectories in complex dynamic environments with the observed history behaviors. It plays a critical role in developing safe human-interactive autonomous systems such as 
self-driving vehicles, social robotics, and intelligent surveillance applications~\cite{alahi2016social,gupta2018social,kosaraju2019social,rudenko2020human}. 

Forecasting the trajectories of humans faces the serious challenges to model the indeterminacy of human behavior.
With the same environment and social interactions, the human might take different plausible actions. For example, to avoid collisions with other pedestrians, one can choose to stop for a moment or speed up.  To generate plausible multi-modal future states instead of a deterministic trajectory, many methods~\cite{gupta2018social,kosaraju2019social,Sadeghian2019CVPR,zhao2019multi} employ the Generative Adversarial Networks~\cite{goodfellow2014generative} (GAN) to spread the distribution of the prediction and cover the space of possible paths, while some methods~\cite{lee2017desire,ivanovic2019trajectron,salzmann2020trajectron++} apply the Conditional Variational Auto-encoder (CVAE)~\cite{sohn2015learning} to explicitly encode the multi-modal distribution with a latent variable. 

\begin{figure}[!t]
\centering
\includegraphics[width=0.43\textwidth]{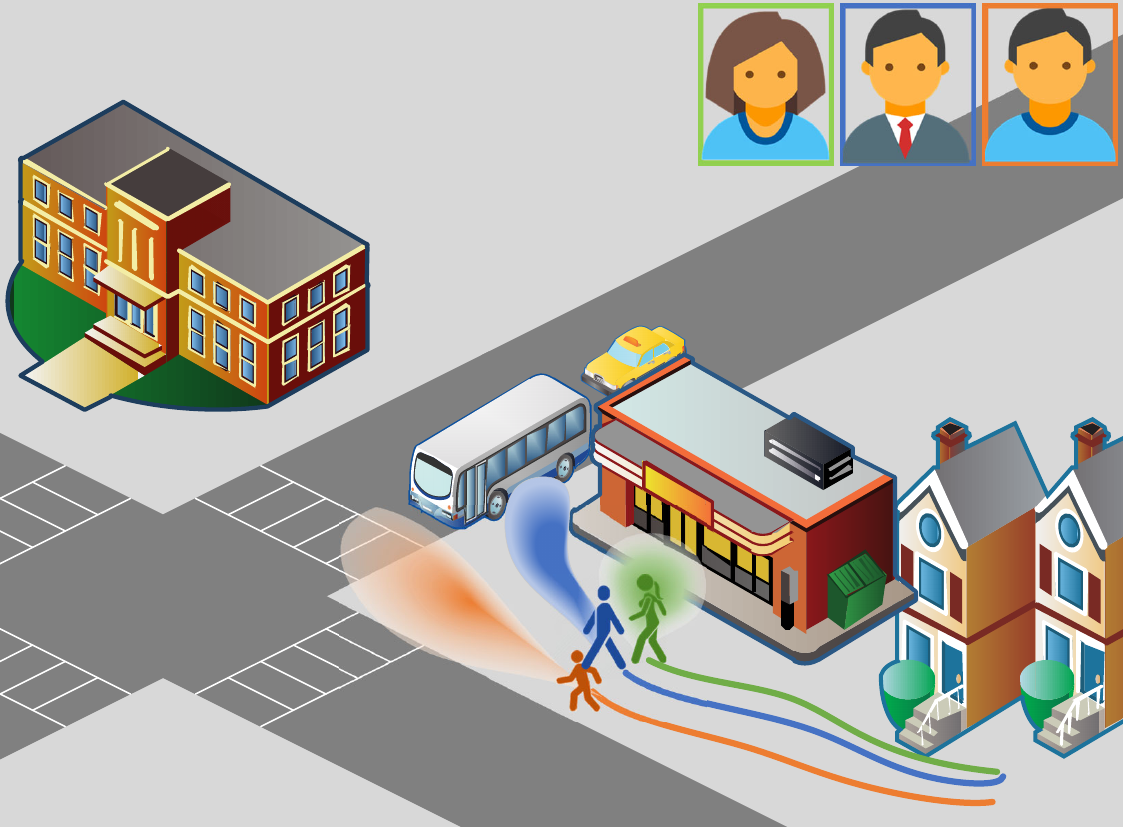}
   \caption{Illustration of the motivation with an example of a family of three. The family members (father, mother, and child) with the similar history trajectories might have different motion patterns due to their habits. (Best viewed in color)  }
\label{fig: top}
\vspace{-0.5cm}
\end{figure}

With the increase of motion diversity and complexity, stochastic prediction with the prior Gaussian distribution is insufficient to cover the wide spectrum of future possibilities. As shown in Figure~\ref{fig: top}, considering a family of three with different habits, the motion patterns are always different even given the similar trajectories. For example, the child tends to go straight to school, while the father might turn right for the bus.
Although some methods\cite{felsen2018will, ivanovic2019trajectron, salzmann2020trajectron++} attempt to use different latent distributions for different persons, discriminative ability of these latent distributions are limited. The model can't always generate appropriate outputs via a most-likely manner. A piece of obvious evidence is the large performance gap in existing generative methods\cite{felsen2018will, salzmann2020trajectron++} between most likely single output and the posterior selected best output. The poor performance of the most likely single output indicates that the learned distribution cannot represent the personalized tendency. The self-driving system samples possible trajectories from learned distribution to make decisions. A good latent distribution is of critical importance for reducing the cost of sampling with no performance drop.

To overcome the above issues, in this paper, we propose a distribution discrimination method (DisDis) to learn the personalized multi-modal distribution, where the behavior pattern of each person is modeled as the latent distribution. Different from other unified prior distribution based generative prediction methods, we learn a distribution discriminator to distinguish the potential behavior patterns. Without any extra supervisory signals, we optimize the distribution discriminator in a self-supervised manner, which encourages the latent variable distributions of the same motion pattern to be similar while pushing the ones of the different motion patterns away. However, blindly increasing the discriminative abilities of latent distributions might break the accuracy of prediction. Therefore, we also optimize the latent distribution with the prediction accuracy via the policy gradient algorithm. To evaluate whether the learned latent variable can represent motion pattern, we further propose a new evaluation metric for stochastic trajectory prediction methods. We calculate the probability cumulative minimum distance (PCMD) curve to comprehensively and stably evaluate the learned model and latent distribution, which cumulatively selects the minimum distance between sampled trajectories and ground-truth trajectories from high probability to low probability. We highlight that DisDis is a plug-and-play module which could be integrated with existing multi-modal stochastic predictive models to enhance the discriminative ability of latent distribution. Note that DisDis method only changes the loss function, which doesn't raise any extra parameters. We show that our DisDis method can achieve improved performance on the ETH~\cite{lerner2007crowds} and UCY~\cite{pellegrini2010improving} datasets.

We summarize the main contributions of this paper as follows:
\begin{itemize}
      \setlength{\itemsep}{2pt}
\setlength{\parsep}{2pt}
\setlength{\parskip}{2pt}
  \item We propose a DisDis method to learn the discriminative personalized latent distribution via the self-supervised contrastive learning and the discrete optimization of consistency constraint.
  \item We further propose a new metric for stochastic methods which comprehensively and stably evaluate predicted trajectories under the multi-modal distribution.
  \item The proposed DisDis method could be integrated with existing stochastic predictors and achieve improved performance in the experiments.
\end{itemize}

\section{Related Work}
\textbf{Social Interactions:}
Many previous methods model the complex social human interaction with energy potentials based on handcrafted rules, e.g. Social Force Model~\cite{helbing1995social}, continuum dynamics~\cite{treuille2006continuum}, Discrete Choice framework~\cite{antonini2006discrete}, Gaussian processes~\cite{tay2008modelling,wang2007gaussian}, crowd analysis~\cite{rodriguez2011data,yamaguchi2011you}, and social sensitivity~\cite{robicquet2016learning}. With the development of deep learning, some methods~\cite{alahi2016social,gupta2018social} use the social pooling layer to aggregate the clues from neighboring trajectories, while attention models~\cite{vemula2018social,Sadeghian2019CVPR,kosaraju2019social,fernando2018soft+,mangalam2020not,liang2019peeking} are also introduced to distinguish the importance of different neighbors or cues. To further analyze social interactions, many recent methods~\cite{zhang2019sr,yu2020spatio,ivanovic2019trajectron,salzmann2020trajectron++,kosaraju2019social,mohamed2020social,huang2019stgat,Sun_2020_CVPR2} employ the spatial temporal graph as the encoder of social interactions. Besides the human interactions, many studies~\cite{lee2017desire,sadeghian2019sophie,xue2018ss,kosaraju2019social} incorporate environment interactions (e.g. physical scene) as additional clues.

\textbf{Deterministic Prediction:}
Most previous methods~\cite{helbing1995social,alahi2016social,lee2016predicting,zhang2019sr,nikhil2018convolutional,xue2018ss,vemula2018social} forecast human trajectory in a deterministic manner. They regard the trajectory forecasting as the sequence prediction problem and apply models like Recurrent Neural Networks (RNN)~\cite{alahi2016social,xue2018ss,vemula2018social,zhang2019sr}, Temporal Convolution Neural Network (TCNN)~\cite{nikhil2018convolutional}, and Inverse Reinforcement Learning (IRL)~\cite{lee2016predicting}. For example, Social LSTM~\cite{alahi2016social} encodes each history trajectory with an LSTM network and connects neighboring trajectories with a social pooling layer, then applies an LSTM decoder for sequential prediction, while Nikhil \emph{et al.}~\cite{nikhil2018convolutional} employ the Temporal CNN to model the temporal connections of human motion and predict future motions. IRL method~\cite{lee2016predicting} regards the motion as a Markov Decision Process for optimization. However, these deterministic
models are infeasible to handle multiple possibilities of human behaviors.

\textbf{Stochastic Prediction:} 
To explore the indeterminacy of future states, many stochastic prediction methods are proposed to predict multiple plausible paths. These methods always incorporate a latent variable into original predictor, such as GAN~\cite{goodfellow2014generative} model or CVAE~\cite{sohn2015learning}. GAN based methods~\cite{gupta2018social,kosaraju2019social,Sadeghian2019CVPR,zhao2019multi,Fang_2020_CVPR,Sun_2020_CVPR} implicitly model the multi-modality and optimize generated trajectories with a discriminator, while CVAE-based methods~\cite{lee2017desire,ivanovic2019trajectron,tang2019multiple,salzmann2020trajectron++,mangalam2020not} explicitly represent the multi-modal distribution and learn it with the constraint between prior and posterior distributions. 
Stochastic methods generate a distribution of potential motion paths, instead of predicting a most-likely single output. 
However, there is a large performance gap in stochastic models\cite{gupta2018social, salzmann2020trajectron++} between the most-likely single output and the best posterior selection from distribution. It indicates that the latent distribution is not well learned to represent personalized future behaviors. To address this problem, we propose a DisDis method, which distinguishes the latent distributions in a self-supervised manner.

\section{Approach}

In this section, we will introduce the proposed distribution discrimination approach. In Section~\ref{sec: problem definition}, we briefly review the human trajectory prediction task and variational prediction methods. We then present our distribution discrimination method to learn personalized latent variable in Section~\ref{sec: dis_dis}. In Section~\ref{sec: discussion}, we discuss the relations with InfoVAE~\cite{zhao2019infovae} and original contrastive learning methods.

\subsection{Problem Definition}
\label{sec: problem definition}

We formulate the pedestrian trajectory prediction task as a sequential prediction problem, which extracts the clues from prior movements and social interactions to predict the possible future navigation movements of pedestrians. The inputs of the prediction model are the observed $N$ history human trajectories within the scene, $\mathbf{x}_i =\{(x_i^t,y_i^t)\in \mathbb{R}^2| t=1,2,\cdots,T_{obs} \} $ for $\forall i \in \{1,2,\cdots,N\} $, where the $(x_i^t,y_i^t)$ is the 2D location at time-instant $t$. Given all above inputs, the goal of our model is to learn the potential distribution to generate plausible future trajectories $\mathbf{y}_i =\{(x_i^t,y_i^t)\in \mathbb{R}^2| t=T_{obs}+1,T_{obs}+2,\cdots.,T_{pred} \} $. 

For the sake of simplicity, we omit the index $i$ in the following descriptions and use $\mathbf{x}, \mathbf{y} $ to respectively denote observed, and future trajectories.

The variational trajectory prediction methods~\cite{lee2017desire,ivanovic2019trajectron,tang2019multiple,salzmann2020trajectron++,mangalam2020not} always learn a latent variable to encode the multi-modal distribution and generate the stochastic future states. The predictive process is as follows: the latent variable $\mathbf{z}$ is generated from a distribution $p_{\theta}(\mathbf{z}|\mathbf{x}) $ and the future trajectory $\mathbf{y} $ is predicted by the generative distribution $g_{\psi}(\mathbf{y}|\mathbf{x},\mathbf{z}) $. This generative process can be formulated as a target distribution $p(\mathbf{y}|\mathbf{x}) $:
 \begin{equation}
  \begin{aligned}
\label{eq: prediction_process} p(\mathbf{y}|\mathbf{x}) = \int g_{\psi}(\mathbf{y}|\mathbf{x},\mathbf{z})p_{\theta}(\mathbf{z}|\mathbf{x})d{\mathbf{z}}.
  \end{aligned}
\end{equation}
In the training process, the distribution $p_{\theta}(\mathbf{z}|\mathbf{x}) $ is constrained with Kullback-Leibler of a prior distribution. Conventionally, the prior distribution is fixed, i.e., a Gaussian distribution $\mathcal{N}(\mathbf{0},\mathbf{I})$. It indicates the predicted trajectories $\mathbf{y}$ is controlled by a fixed $\mathcal{N}(\mathbf{0},\mathbf{I})$, instead of the real-world personalized motion distribution. 

\begin{figure*}[!t]
\centering
\includegraphics[width=0.9\textwidth]{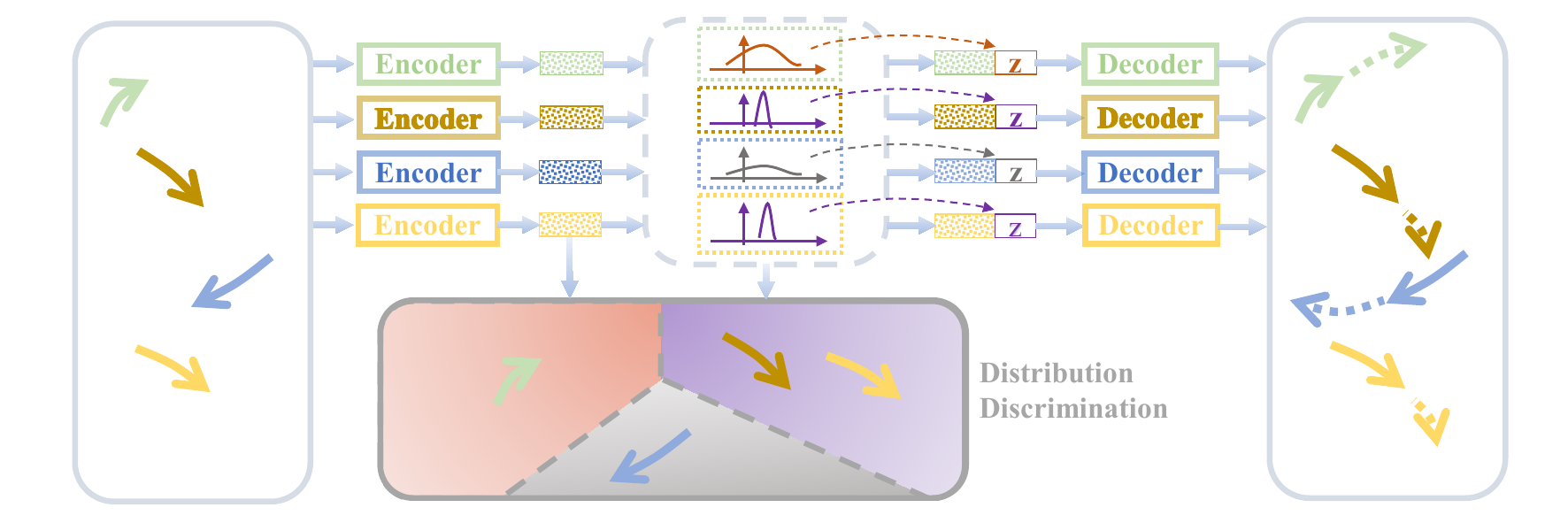}
   \caption{Training process for the DisDis method. DisDis regards the latent distribution as the motion pattern and optimizes the trajectories with the same motion pattern to be close while the ones with different patterns are pushed away, where the same latent distributions are in the same color. For a given history trajectory, DisDis predicts a latent distribution as the motion pattern, and takes the latent distribution as the discrimination to jointly optimize the embeddings of trajectories and latent distributions. (Best viewed in color.)}
\label{fig: overview}
\vspace{-0.3cm}
\end{figure*}

\subsection{Distribution Discrimination}
\label{sec: dis_dis}

In this work, we present that the latent variable denotes the personalized motion patterns. The core of our work is to learn a discriminative latent variable distribution. Unlike original variational prediction methods which learn the latent variable with only the consistency constraint between proposal distribution and the prior distribution, we learn a distribution discriminator to distinguish the potential behavior patterns. We enforce three important criteria in the optimization process of latent distribution:
\begin{itemize}
    \setlength{\itemsep}{1pt}
\setlength{\parsep}{1pt}
\setlength{\parskip}{1pt}
  \item The latent variable distribution should help to predict the accurate pedestrian trajectories. Besides the diversity of multi-modal motions, the accuracy of the prediction model is also of critical importance in the trajectory forecasting system.
  \item The latent variable distribution should be consistent for the history trajectories and corresponding future predictions. We assume that the motion pattern of the same person is consistent due to his/her habit and character. 
  \item The latent variable distributions should be discriminative. To model the personalized motion patterns, we encourage the latent variable distributions of the same motion pattern to be similar while pushing the ones of the different motion patterns away.  
\end{itemize} 

To achieve these criteria, we formulate the following loss function to learn the latent variable distribution:
 \begin{equation}
  \begin{aligned}
\label{eq: our_loss} \mathcal{L}_{DisDis}(\psi,\phi,\theta) &=  \mathcal{L}_{1}(\phi,\theta)  + \lambda \mathcal{L}_{2}(\phi,\psi)  +\mu \mathcal{L}_{3}(\phi) \\
   & =   \mathcal{L}_{KL}\big(q_{\phi}(\mathbf{z}|\mathbf{x},\mathbf{y})||p_{\theta}(\mathbf{z}|\mathbf{x})\big)\\
   & ~~~~ -\lambda \mathbb{E}_{q_{\phi}(\mathbf{z}|\mathbf{x},\mathbf{y})}[\log g_{\psi}(\mathbf{y}|\mathbf{x},\mathbf{z})]  \\
   & ~~~~ -\mu  \mathbb{E}_{q_{\phi}(\mathbf{z})}\bigg[ \log \frac{h(\mathbf{z},\mathbf{x})}{\sum_{  q_{\phi}(\mathbf{z}) }h(\mathbf{z},\mathbf{x})}\bigg],                
  \end{aligned}
\end{equation}

where $ h(\mathbf{z},\mathbf{x}) $ denotes the density ratio which preserves the mutual information between $ \mathbf{z} $ and $ \mathbf{x}$, $\lambda$ and $\mu $ are two hyper-parameters to balance the effects of different terms in the objective function for a good trade-off. The $q_{\phi}(\mathbf{z}) $ is the marginal distribution evaluated with unbiased samples from the distribution $q_{\phi}(\mathbf{z})= q_{\phi}(\mathbf{z}|\mathbf{x})p_{D}(\mathbf{x}) $~\cite{zhao2019infovae}, where $p_{\mathcal{D}}(\mathbf{x}) $ denotes the true underlying distribution approximated by a training set (or the samples selected in the batch). We explain each term of our DisDis loss function as follows:
\begin{itemize}
   \setlength{\itemsep}{2pt}
\setlength{\parsep}{2pt}
\setlength{\parskip}{2pt}
  \item Inspired by CVAE~\cite{sohn2015learning}, we reduce the KL divergence between the proposal distribution $q_{\phi}(\mathbf{z}|\mathbf{x},\mathbf{y}) $ and prior distribution $p_{\theta}(\mathbf{z}|\mathbf{x})$ in the first term of~\eqref{eq: our_loss}. The proposal distribution encodes the future motions, while the prior distribution only observes the history trajectories. This negative KL divergence encourages the consistency of latent variables generated from history and future motions.

  \item We also encourage the learned latent variable to predict the real future trajectory. Different from the reconstruction part of conventional VAE methods which sample latent variables from distribution to optimize the predictor $g_{\psi}$ independently, we jointly learn the latent variable distribution and the predictor. We apply the discrete optimization algorithms, e.g., policy gradient, to learn the latent variable distribution, since the sampling process can not be directly optimized with the gradient back-propagation. 
 \begin{equation}
  \begin{aligned}
\label{eq: policy_gradient} \nabla \mathcal{L}_{2}(\phi) 
                               & \approx -\frac{1}{N}\sum_{i=1}^{N}\big[\nabla_{\phi} \log q_{\phi}(\mathbf{z}_i|\mathbf{x}_i,\mathbf{y}_i) R \big],
  \end{aligned}
\end{equation}
where the policy is defined as the sampling process of the latent variable, and the reward $R$ denotes the negative $\mathcal{L}_{1} $. The policy gradients are estimated with the Monte Carlo method with selected $N$ samples.
  
  \item The last term of~\eqref{eq: our_loss} optimizes the discriminative ability of latent variables. The main intuition behind the distribution discrimination is to distinguish personalized motion patterns (the latent variable). As shown in Figure~\ref{fig: overview}, we optimize a discriminative latent space, where the trajectories with the same motion pattern have similar representation while negative trajectories are pushed away. We predict a latent distribution as the motion pattern to optimize the embeddings of trajectories and latent distributions as pseudo labels. Note that profiting from the consistency constraint in $\mathcal{L}_{2} $, we optimize prior distribution $p_{\theta}(\mathbf{z}|\mathbf{x})$ to replace $q_{\phi}(\mathbf{z}|\mathbf{x}) $, (from  $\mathcal{L}_{3}(\phi) $ to $\mathcal{L}_{3}(\theta) $ ). In the following, we give two perspectives to understand our distribution discrimination formulation in $\mathcal{L}_{3}(\theta) $, including the contrastive metric learning and mutual information optimization. 
\end{itemize}

\emph{The perspective of contrastive metric learning:} We encourage the model to learn a discriminative latent variable space, where the latent distributions (motion patterns) denote a recognition goal of trajectories. Considering the embedding $\mathbf{f}$ of the trajectory $\mathbf{x}$, we define the $h(\mathbf{z},\mathbf{x})$ as an energy-based model:
 \begin{equation}
  \begin{aligned}
\label{eq: loss3_h} h(\mathbf{z},\mathbf{x}) = \exp(\mathbf{z}^TW^T\mathbf{f}),
  \end{aligned}
\end{equation}
where $W^T\mathbf{f} $ is a linear transformation to predict the motion pattern $\mathbf{z} $ with input $\mathbf{x}$. We assume that the latent variable denotes the motion pattern, and call the trajectories with same motion pattern as positive trajectories $\mathbf{f}^+$ while the others as negative ones $\mathbf{f}^-$. When using this linear transformation to parameterize the $p_{\theta}(\mathbf{z}|\mathbf{x})$, we can obtain that $\mathbf{z} = W^T\mathbf{f} = W^T\mathbf{f}^+ $, and write the original $\mathcal{L}_{3}(\theta)$ as:
 \begin{equation}
  \begin{aligned}
\label{eq: loss3_dis} \mathcal{L}_{3}(\theta) &= - \log \frac{\exp( d(\mathbf{f},\mathbf{f}^+)  )}{
\sum_{\mathbf{f}_i \sim p_{\mathcal{D}}(\mathbf{f})}      \exp( d(\mathbf{f},\mathbf{f}_i)  )}   \\
&= \log \big(1+  \sum_{\mathbf{f}_i \sim p_{\mathcal{D}}(\mathbf{f})} \exp(d(\mathbf{f},\mathbf{f}_i) - d(\mathbf{f},\mathbf{f}^+)  )  \big),  
  \end{aligned}
\end{equation}
where, $ d(\mathbf{f},\mathbf{f}^+) = \mathbf{f}^TWW^T\mathbf{f}^+ $ is the Mahalanobis distance of embedding $\mathbf{f}$, which can be replaced with other distances, i.e. cosine distance. We represent the latent variable $\mathbf{z}$ with a projection of the trajectory embedding or the positive one. This formulation is regarded as a typical metric learning objective which focuses entirely on shortening the distance between the embeddings of the trajectories with the same motion pattern while enlarging the distances between those of different motion patterns.

\emph{The perspective of mutual information optimization:}
The distribution discrimination is also equal to optimizing the mutual information $I(\mathbf{z},\mathbf{x}) $, which denotes the dependence between observed trajectories $\mathbf{x}$ and corresponding latent motion pattern $\mathbf{z}$. When the energy-based function $h(\mathbf{z},\mathbf{x}) $ is defined as the density ratio between $p_{\theta}(\mathbf{z}|\mathbf{x}) $ and $ p_{\theta}(\mathbf{z}) $:
 \begin{equation}
  \begin{aligned}
\label{eq: loss3_h2} h(\mathbf{z},\mathbf{x}) \propto \frac{p_{\theta}(\mathbf{z}|\mathbf{x})}{p_{\theta}(\mathbf{z}) },
  \end{aligned}
\end{equation}
we reformulate the $\mathcal{L}_{3} $ as a lower bound of the mutual information as:
 \begin{equation}
  \begin{aligned}
\label{eq: loss3_mutual} \mathcal{L}_{3}(\theta) &= -\mathbb{E}_{\mathbf{z}} \log \bigg[ \frac{ \frac{p_{\theta}(\mathbf{z}|\mathbf{x})}{p_{\theta}(\mathbf{z})}} {  \frac{ p_{\theta}(\mathbf{z}|\mathbf{x})}{ p_{\theta}(\mathbf{z})}  + \sum_{\mathbf{z}_i \neq \mathbf{z}} \frac{p_{\theta}(\mathbf{z}_i|\mathbf{x})}{p_{\theta}(\mathbf{z}_i)}     }\bigg]   \\
      & \geq -I(\mathbf{z},\mathbf{x}) +\log(N'),
  \end{aligned}
\end{equation}
where $N' $ is the number of the selected negative samples. Thus minimizing the distribution discrimination loss is equal to increasing the mutual information between $\mathbf{x} $ and $\mathbf{z}$, which reduces the information
preference problem~\cite{chen2016variational}. The formal proof is given by~\cite{oord2018representation}, which we recapitulate for our variational prediction framework in the appendix. 

\subsection{Discussion}
\label{sec: discussion}

\textbf{Relation to InfoVAE}: The InfoVAE~\cite{zhao2019infovae} method optimizes the mutual information by approximating $I(\mathbf{z},\mathbf{x}) $ with the KL divergence between $q_{\phi}(\mathbf{z}) $ and $p(\mathbf{z}) $. However, evaluating $q_{\phi}(\mathbf{z}) $ with the sampling of $\mathbf{x} \sim p(\mathbf{x}) $ might introduce extra noises, which influence the calculation of the KL divergence. In fact, we are primarily interested in maximizing mutual information instead of its precise value. Thus, in our DisDis method, we optimize the lower bound of mutual information instead of calculating KL divergences. In the experiments, we maintain the mutual information term in InfoVAE, and add our contrastive discrimination. The results show the advantages of adding distribution discrimination.

\textbf{Relation to Contrastive Learning}: Contrastive learning~\cite{He_2020_CVPR,tian2019contrastive,wu2018unsupervised} is a kind of self-supervised methods to encourage the close embeddings under different views of the same scene. There are two main differences between conventional contrastive learning methods and our DisDis method: 1) the goal of our method is to learn the latent variable distribution instead of unsupervised representation learning; 2) we encourage the trajectory embeddings to be close to the ones of the same motion pattern, instead of different views of the same trajectory. An intuitive explanation is that we argue the latent variable as pseudo labels while conventional contrastive learning methods regard the augmented samples of same trajectory as a class.

\section{Experiment}

In this section, we first propose a new evaluation metric for stochastic prediction algorithms, which considers the comprehensive evaluation under the learned distribution. Then, we evaluate our DisDis method on two publicly available human trajectory prediction datasets: ETH~\cite{pellegrini2010improving} and UCY~\cite{lerner2007crowds}. Quantitatively, we conduct the ablation studies to compare our method with baseline models including VAE, CVAE, InfoVAE and the combination of VAE and contrastive learning. Besides, we also compare our method with other state-of-the-art human trajectory prediction approaches. Qualitatively, we provide the visualization explanation under the real world environment and show that the proposed DisDis method can learn the discriminative latent pattern models.

\begin{figure}[t]
\centering
\subfigure[ADE-based CDM curves]{
\includegraphics[scale=0.5]{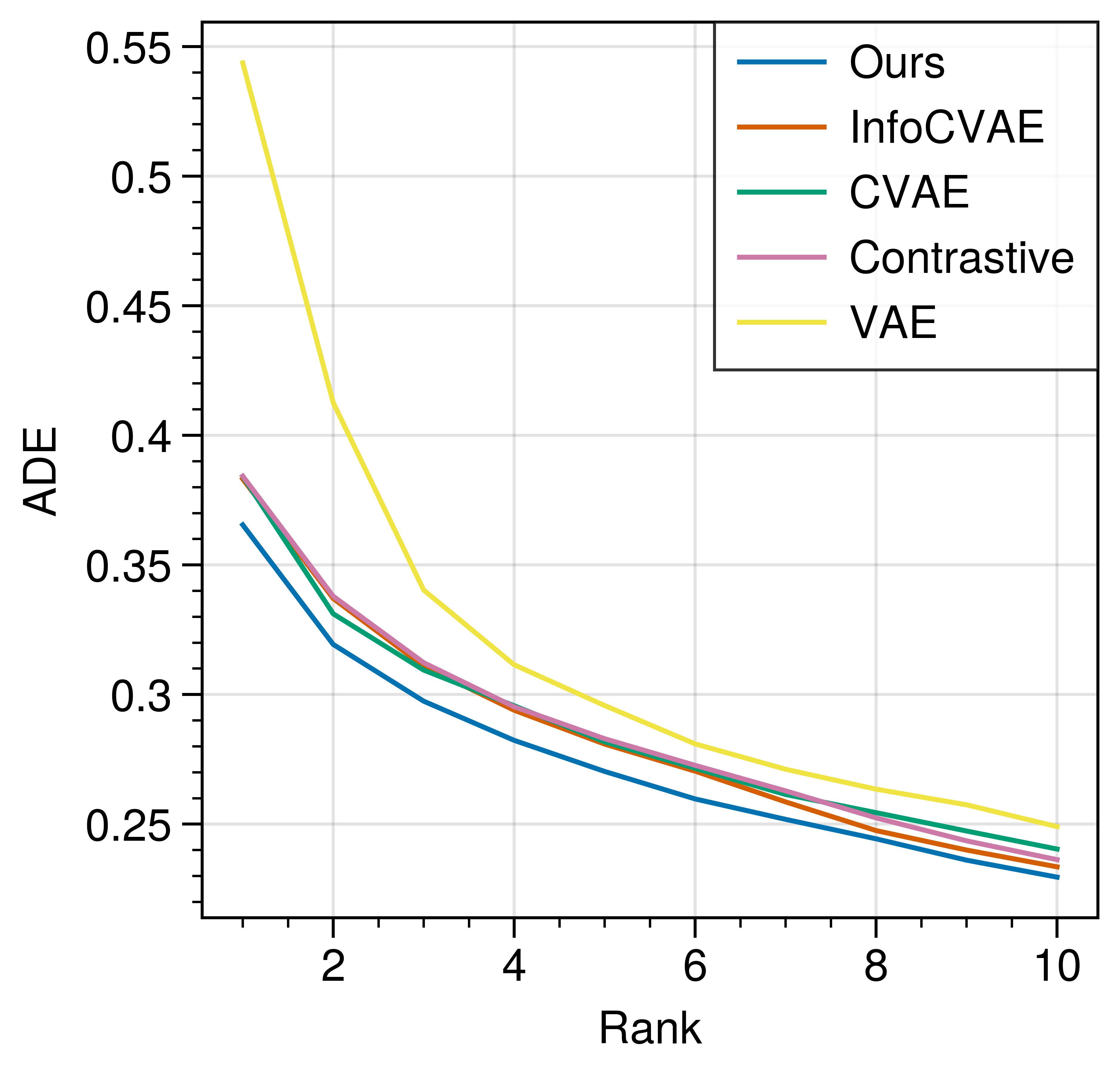}
}
\subfigure[FDE-based CDM curves]{
\includegraphics[scale=0.5]{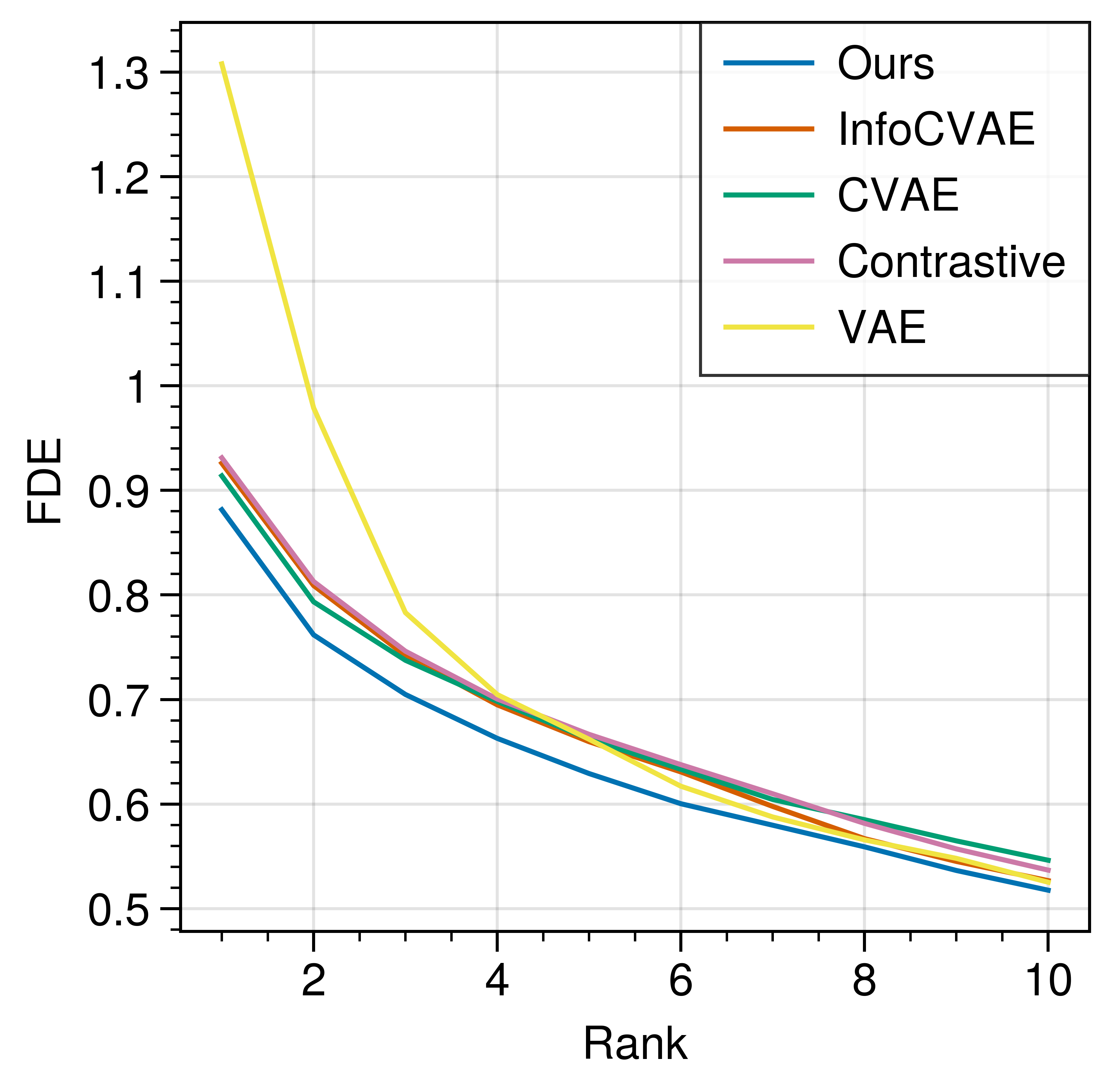}
}
\caption{Ablation comparisons with baseline methods on ADE-based and FDE-based CDM curves. We set sampling $M=80$ and plot top 10 ranks. The lower is the better.} 
\label{fig:baseline}
\vspace{-0.2cm}
\end{figure}

\subsection{Datasets and Experimental Settings}

\textbf{Datasets:} Our experiments are conducted on two publicly available datasets: ETH~\cite{pellegrini2010improving} and UCY~\cite{lerner2007crowds}, which serve as the major benchmark for human trajectory prediction task. These datasets contain 1536 detected pedestrians in five unique scenes: Zara1, Zara2, Univ, ETH, and Hotel. We follow the commonly used leave one set out cross-validation evaluation strategy, i.e., training on four scenes and testing on the remaining one~\cite{gupta2018social,kosaraju2019social,huang2019stgat,salzmann2020trajectron++}. All trajectories are sampled in 0.4 seconds (20 frames), where the first 3.2 seconds correspond to observed data and the next 4.8 seconds correspond to predicted future data.

\begin{table*}
\caption{Comparison with several state-of-the-art models with PCMD curves. All the models are reproduced with the code previously published online. We show the values on PCMD curves at rank $1$, $5$ and $20$. The lower is the better.}
\renewcommand\arraystretch{1.3}
\renewcommand\tabcolsep{2pt}
\begin{center}
\begin{tabular}{l|c|c|c|c|c|c}
\hline
\multirow{2}*{Methods}  &  \multicolumn{6}{c}{$PCMD_{ADE}@\{\frac{1}{M}/ \frac{5}{M} / \frac{20}{M} \} \vert M=80$  } \\
\cline{2-7}
~ & ETH & HOTEL &  ZARA1&  ZARA2 &UNIV & AVG \\
\hline
Social-GAN~\cite{gupta2018social} & 0.98/0.82/0.73  & 0.63/0.54/0.48 & 0.47/0.38/0.33 & 0.39/0.33/0.3 & 0.64/0.58/0.55  & 0.62/0.53/0.48 \\
STGAT~\cite{huang2019stgat}  & 1.03/0.89/0.78 & 0.59/0.45/0.38  & 0.53/0.39/0.33  & 0.43/0.34/0.29  & 0.68/0.59/0.56  & 0.65/0.53/0.47 \\
Social-STGCNN~\cite{mohamed2020social}   & 1.01/0.87/0.76 & 0.74/0.53/0.42  & 0.57/0.42/0.34  & 0.51/0.38/0.31  & 0.71/0.59/0.51  & 0.71/0.56 0.47 \\
Trajectron++~\cite{salzmann2020trajectron++}  & 0.73/0.58/0.43  & 0.27/0.19/\textbf{0.11}  & 0.30/\textbf{0.20}/\textbf{0.13}  & \textbf{0.22}/\textbf{0.16}/\textbf{0.10}  & 0.39/\textbf{0.27}/\textbf{0.17}  & 0.38/0.28/0.19 \\
Trajectron++*~\cite{salzmann2020trajectron++}  & 1.02/0.86/0.65  & 0.52/0.24/0.15  & 0.43/0.31/0.19  & 0.32/0.25/0.15  & 0.52/0.41/0.25  & 0.56/0.41/0.28 \\
\hline
DisDis(Ours) & \textbf{0.71}/\textbf{0.55}/\textbf{0.38} & \textbf{0.25}/\textbf{0.17}/\textbf{0.11}  & \textbf{0.28}/\textbf{0.20}/\textbf{0.13}  & \textbf{0.22}/\textbf{0.16}/\textbf{0.10}  & \textbf{0.36}/\textbf{0.27}/\textbf{0.17}  & \textbf{0.36}/\textbf{0.27}/\textbf{0.18} \\
DisDis* & 1.02/0.83/0.61 & 0.41/0.22/0.15  & 0.42/0.30/0.19  & 0.32/0.25/0.15  & 0.53/0.41/0.25  & 0.54/0.40/0.27 \\
\hline\hline
\multirow{2}*{Methods}  &  \multicolumn{6}{c}{$PCMD_{FDE}@\{\frac{1}{M}/ \frac{5}{M} / \frac{20}{M} \} \vert M=80$  } \\
\cline{2-7}
~ & ETH & HOTEL &  ZARA1&  ZARA2 &UNIV & AVG \\
\hline
Social-GAN~\cite{gupta2018social}  & 1.98/1.58/1.40 & 1.36/1.14/1.02  & 1.02/0.79/0.67  & 0.87/0.70/0.63  & 1.38/1.24/1.17  & 1.32/1.09/1.03 \\
STGAT~\cite{huang2019stgat}  & 2.20/1.86/1.52 & 1.21/0.89/0.73  & 1.17/0.81/0.67  & 0.94/0.70/0.60  & 1.49/1.28/1.20  & 1.40/1.11/0.94 \\
Social-STGCNN~\cite{mohamed2020social}  & 1.83/1.54/1.29  & 1.42/1.00/0.74  & 1.13/0.79/0.57  & 0.97/0.69/0.52  & 1.38/1.15/0.98  & 1.35/1.03/0.82 \\
Trajectron++~\cite{salzmann2020trajectron++}  & 1.73/1.36/0.92  & 0.57/0.37/0.18  & 0.76/\textbf{0.49}/0.27  & 0.57/\textbf{0.41}/0.23  & 0.99/0.67/\textbf{0.36}  & 0.93/0.66 /0.39 \\
Trajectron++*~\cite{salzmann2020trajectron++}  & 2.12/1.77/1.24  & 1.01/0.43/0.24  & 0.96/0.65/0.35  & 0.72/0.54/0.28  & 1.16/0.88/0.49  & 1.19/0.85/0.52 \\
\hline
DisDis(Ours) & \textbf{1.67}/\textbf{1.24}/\textbf{0.75} & \textbf{0.52}/\textbf{0.33}/\textbf{0.18}  & \textbf{0.73}/0.50/\textbf{0.27}  & \textbf{0.56}/0.41/\textbf{0.22}  & \textbf{0.92}/\textbf{0.66}/0.37    & \textbf{0.88}/\textbf{0.63}/\textbf{0.36} \\
DisDis* & 2.13/1.68/1.21 & 0.75/0.39/0.22  &0.93/0.64/0.35  & 0.70/0.54/0.30  & 1.17/0.86/0.49    & 1.14/0.82/0.51 \\
\hline
\end{tabular}
\end{center}
\label{table:sota}
\vspace{-0.4cm}
\end{table*}

\textbf{Evaluation Metric:} The same evaluation metrics as prior methods~\cite{gupta2018social,kosaraju2019social,huang2019stgat,salzmann2020trajectron++} are adopted, including Average Displacement Error (ADE) and Final Displacement Error (FDE). ADE computes the mean square error (MSE) overall estimated positions in the predicted trajectory and ground-truth trajectory, while FDE computes the distance between the predicted final destination and the ground-truth final destination. 

Instead of the deterministic prediction, stochastic predicted methods generate the future motions under a multi-modal distribution. 
There are two widely used strategies to compute the ADE and FDE for the predicted multi-modal trajectories: 
\begin{itemize}
  \item \textbf{``Most-likely'' strategy} predicts the most-likely single output under the multi-modal distribution and uses it to calculate the ADE and FDE. It changes the stochastic prediction to deterministic one to obtain the most-likely performance.
  \item \textbf{``Best-of-N'' strategy} generates N samples (i.e. $N=20$~\cite{gupta2018social,salzmann2020trajectron++,kosaraju2019social}) based on the predicted distribution and selects the closest sample to the ground truth for computing ADE and FDE metrics. It evaluates the highest performance of the model by considering as many samples as possible.
\end{itemize} 
However, both two strategies ignore the evaluation of learned latent distribution. It is of critical importance for stochastic prediction algorithms since the latent variable denotes the personalized motion pattern of human. In this paper, we propose a new evaluation strategy, called probability cumulative minimum distance (PCMD) curve, to evaluate the prediction model under the multi-modal distribution. There are two motivations of our PCMD evaluation metric: 1) Though the “Most-likely” strategy considers the latent distribution, it only evaluates the most-likely point. The single point can’t represent the latent distribution. 2) “Best-of-N” strategy samples multiple points from the latent distribution. However, it only uses the best one for evaluation, which ignores the probability of each sample.  Different from these two strategies, PCMD considers the predictions with corresponding probabilities, which evaluate the model under the whole latent distribution. Inspired by the Cumulative Match Characteristic (CMC) curve in the recognition tasks, we cumulatively calculate the minimum ADE and FDE of sampled trajectories from high probability to low probability.

\begin{table}[!t]
\centering
\caption{Comparison with SOTA models with Best-of-20 strategy}
\label{table: bestof20}
\begin{center}
\begin{tabular}{c|c|c}
    \hline
    Methods& ADE & FDE \cr
    \hline
    Social-GAN~\cite{gupta2018social} & 0.58 & 1.18  \cr
    STGAT~\cite{huang2019stgat} & 0.43 & 0.83 \cr
    Social-STGCNN~\cite{mohamed2020social} & 0.44 & 0.75 \cr
    Trajectron++~\cite{salzmann2020trajectron++} & 0.19 & 0.41 \cr
    Trajectron++*~\cite{salzmann2020trajectron++} & 0.32 & 0.55 \cr
    \hline
    DisDis(Ours) & \textbf{0.17} & \textbf{0.37} \cr
    DisDis* & 0.26 & 0.47 \cr
    \hline
\end{tabular}
\end{center}
\vspace{-0.5cm}
\end{table}

Formally, given the distribution $p_\theta(\mathbf{z}\vert \mathbf{x}), \mathbf{z}\in\Omega$, we can define a projection as
\begin{equation}
F(\tau) = \mathbb{E}_{\mathbf{z}\in \Omega} \mathbb{I}(p_\theta(\mathbf{z}\vert \mathbf{x}) \geq \tau)
\end{equation}
where the input denotes a selected probability value $\tau \in (\min p_\theta(\mathbf{z}\vert \mathbf{x}), \max p_\theta(\mathbf{z}\vert \mathbf{x}) ) $, $ \mathbb{I} $ is an indicator function, while the output $F(\tau) \in(0,\Vert\Omega \Vert) $ denotes the interval length of $\mathbf{z}$ satisfying conditions. We define $k= \frac{F(\tau)}{\Vert\Omega \Vert}$ to normalize it into $(0,1) $, if the latent space $\Omega $ is finite. (The condition $\Vert\Omega \Vert \to \infty $ can be ignored when we discretely approximate it by the Monte Carlo methods. ) We obtain the values of PCMD curve as:
\begin{equation}
PCMD(k) =\mathbb{E}_{\mathbf{x}} \min\{ \mathcal{D}(\mathbf{z}) \vert p_\theta(\mathbf{z}\vert \mathbf{x}) \geq \tau,\mathbf{z} \in \Omega\},
\end{equation}
where $ \mathcal{D}(\mathbf{z})$ denotes the ADE or FDE distance between the ground-truth and the sampled predicted trajectories based on $\mathbf{z}$. The $PCMD(\mathbf{k}) $ denotes the minimum ADE or FDE distance from all sampled latent variable satisfying condition $p_\theta(\mathbf{z}\vert \mathbf{x}) \geq \tau $. The lower PCMD value indicates we can obtain better performance with the same sampling number. It is critical for auto-driving system since more sampling means more time-delay. 

\begin{figure*}[!t]
\centering
\includegraphics[width=0.98\textwidth]{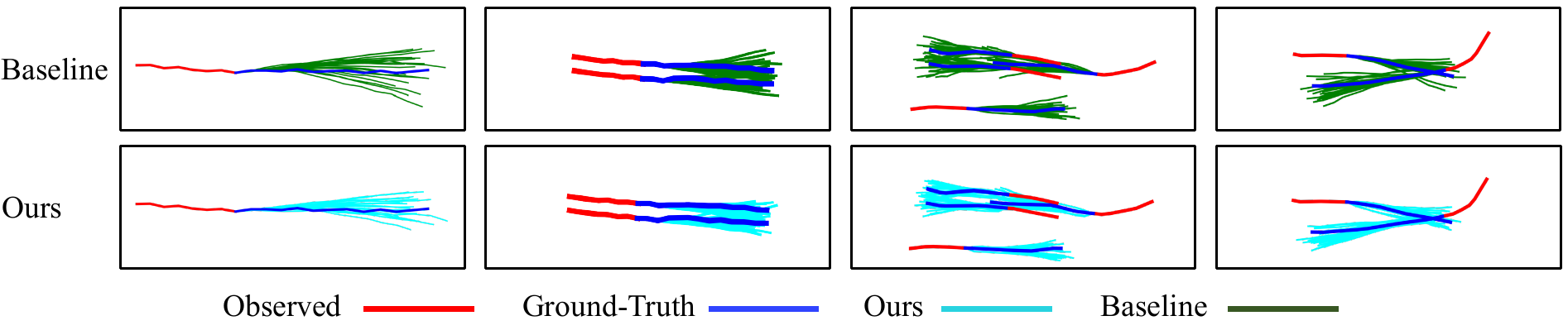}
    \caption{Illustration of the learned distributions. We compare the learned distributions of our method and Trajectron++~\cite{salzmann2020trajectron++} baseline on 4 different scenarios including a pedestrian walking alone; pedestrians walking in parallel; pedestrians following the people ahead; and pedestrians meeting from different directions. We plot 20 top rank trajectories to show the distributions.}
\label{fig: results_distribution}
\vspace{-0.2cm}
\end{figure*}

To numerically calculate the PCMD, we sample M variables $ Z= \{\mathbf{z}_i \in \Omega \vert i=1,2,\cdots,M \} $ and sort these variables with the probability $p_\theta(\mathbf{z}_i\vert \mathbf{x})$ from large to small to obtain $Z_{sort} = \{\mathbf{z}^*_i \} $. Then, the values of PCMD are calculated as:
\begin{equation}
PCMD(k)  \!=\! \mathbb{E}_{\mathbf{x}} \min\{\mathcal{D}(\mathbf{z}) \vert z \!\in\! \{ z^*_1,z^*_2, \cdots, z^*_m\}\},
\end{equation}
where $k = \frac{m}{M}$ denotes the ranking rate in the numerical calculation, e.g. we calculate the minimum ADE/FDE of top 20 probability trajectories, when $k = \frac{20}{M} $. Please see the detailed derivations in the appendix.

The proposed PCMD curve has two significant advantages over existing evaluation metrics: more comprehensive and more stable. First, PCMD comprehensively analyzes the latent distribution by the accumulation of probability, while other evaluation metrics can be regarded as a part of the PCMD curve. $PCMD(1/M)$ denotes the most-likely single output ADE/FDE performance, while $PCMD(N/M)$ denotes the best ADE/FDE performance of trajectories with top N probability. Note that in the experiments we used the same sampling number for all methods for fair comparison. 

Besides, the proposed PCMD curve is more stable than the ``Best-of-N'' evaluation since random sampling brings large performance jitter due to the randomness. PCMD reduces this randomness by the larger sampling $M$ ($M\gg N$) and more stable expression (a curve rather than a number). For the discrete latent distribution, we can obtain a totally fixed evaluation due to the complete sampling. Sampling $M$ trajectories is only conducted when we evaluate different models and select the better one. In a real-life scenario, we can only generate a part of trajectories with higher probabilities using the selected model.

\subsection{Implementation Details}
As a plug-and-play module, our DisDis method can be integrated to train existing stochastic predictors. To evaluate the effectiveness of the learned latent distribution of our method, we select the SOTA Trajectron++~\cite{salzmann2020trajectron++} $\footnote{We reproduced the Trajectron++~\cite{salzmann2020trajectron++} method with the official released code from \url{https://https://github.com/StanfordASL/Trajectron-plus-plus}. }$ method, which is based on CVAE and InfoVAE, as the baseline and incorporate our distribution discrimination. We follow the encoder and decoder networks of Trajectron++, and make some modifications for convenient and fair comparison: 1) we did not use the scene clues (e.g. local map) and removed complex dynamics integration to obtain simpler embedding; 2) we modeled the latent variable with a higher dimensional discrete variable ($|Z|=80$) which better evaluates the learned latent variable distribution; 3) we used a fixed most likely output to replace the Gaussian output structure, which guarantees that the diversity is only from the learned latent distribution. We adopt the same data preprocessing strategies with Trajectron++~\cite{salzmann2020trajectron++}, which rotates trajectories in a scene around the scene's origin in a interval of $15^{\circ}$. Due to the same networks, the ultimate performances of baseline and our method is similar. The main difference is DisDis learning a more discriminative latent distribution.

\subsection{Quantitative Evaluation}

\textbf{Comparison with Baseline Methods:} 
We conduct ablation study to compare our method with the following baseline methods with the same network implementation and different latent distribution learning: 1) VAE~\cite{kingma2014auto} optimizes the latent distribution towards a fixed prior distribution. 2) CVAE~\cite{sohn2015learning} learns the latent variable with the KL divergence between the proposal distribution $q_{\phi}(\mathbf{z}|\mathbf{x},\mathbf{y}) $ and prior distribution $p_{\theta}(\mathbf{z}|\mathbf{x}) $. 3) CVAE + InfoVAE~\cite{zhao2019infovae} further considers the mutual information between $ \mathbf{x} $ and $\mathbf{z} $ by approximating the distribution $q_{\phi}(\mathbf{z}) $. 4) CVAE + Contrastive Learning optimizes to the similar latent variable for different views of the same trajectory. 

\begin{figure}[!t]
\centering
\includegraphics[width=0.48\textwidth]{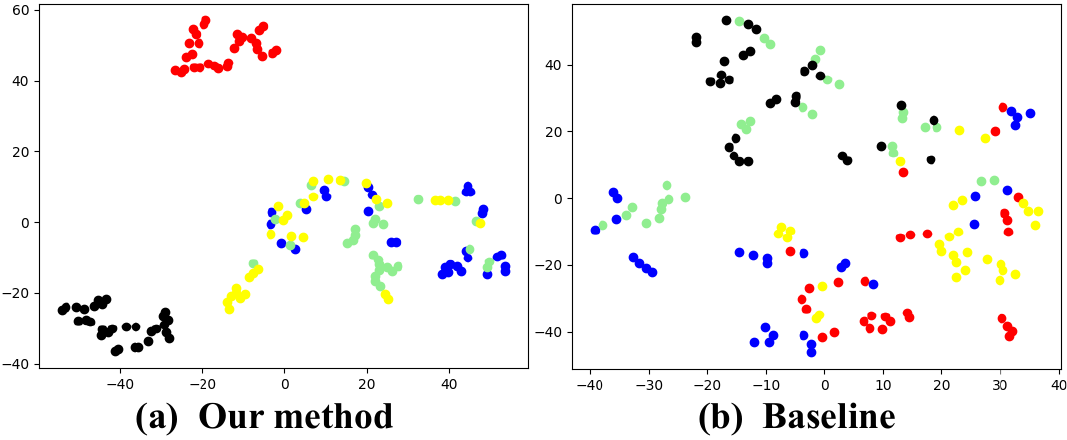}
   \caption{The t-SNE visualization of latent variables learned by DisDis and Trajectron++. Best viewed in color. }
\label{fig: tsne}
\vspace{-0.5cm}
\end{figure}

\begin{figure*}[!t]
\centering
\includegraphics[width=0.95\textwidth]{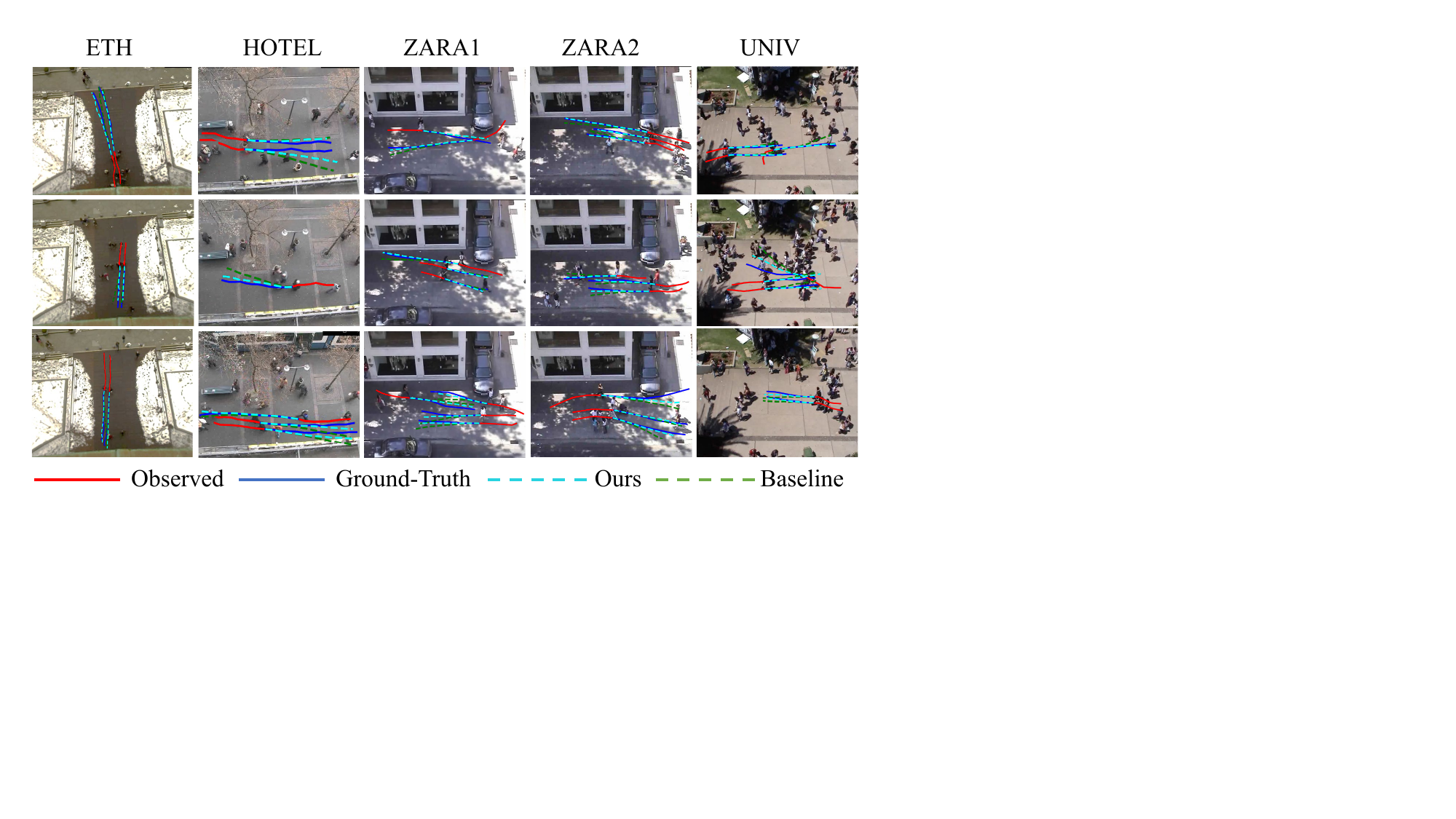}
   \caption{ Visualization examples of our DisDis method and baseline Trajectron++~\cite{salzmann2020trajectron++} method in the different scenes in the both ETH and UCY datasets. We plot the most likely predictions as the results, which quantitatively demonstrate the effectiveness of distribution discrimination to learn the discriminative latent distribution. (Best viewed in color.)}
   \label{fig: results_examples}
\vspace{-0.2cm}
\end{figure*}

We evaluate our method and the baselines in our proposed PCMD curve metric. As shown in Figure~\ref{fig:baseline}, the original VAE baseline obtains the worst performance on top probability ranks, which indicates bias between 
the fixed prior distribution and the real-world personalized motion distribution, while our DisDis obtains best top rank performance due to the more discriminative distribution. Compared with CVAE and Info+CVAE, our DisDis also achieves an improved performance, which demonstrates optimizing the mutual information by distribution discrimination is a better manner than approximately calculating the  $q_{\phi}(\mathbf{z}) $. Besides, the DisDis method outperforms directly applying contrastive self supervised learning on VAE. It demonstrates that using the latent distributions to represent motion patterns is better than regarding each trajectory as one motion pattern. In our assumption, the trajectories with the same motion pattern are clustered in the embedding space. When increasing the number of cumulative probabilities, performance gaps among all methods become small, which is because we use the same encoder and decoder networks for all baseline methods and our method. \emph{Note that} the performances of all methods will converge to the same point when we sample enough trajectories since we use the same networks. The main difference among these methods is the optimization strategy of the latent variables. In this setting, the latent distribution of our DisDis method outperforms the ones of other methods since we can achieve same performance with fewer sampling.

\textbf{Comparison with SOTA Stochastic Methods:} We always compare our method with a wide range of SOTA stochastic methods, including: Social-GAN~\cite{gupta2018social}, STGAT~\cite{huang2019stgat} ,Social-STGCNN~\cite{mohamed2020social}, and Trajectron++~\cite{salzmann2020trajectron++}. All the models are reproduced with the code previously published online, and \textbf{fairly} evaluated with PCMD curve metric. As shown in Table~\ref{table:sota}, we provide some detailed points including the performance of $PCMD@\{\frac{1}{M},\frac{5}{M},\frac{20}{M}\}$. For convenience, we define the $PCMD@\{\frac{1}{M},\frac{5}{M},\frac{20}{M}\}$ as Rank $1$, Rank $5$ and Rank $20$. Rank $1$ of ADE/FDE denotes the most-likely single output ADE/FDE performance, while Rank $20$ ADE/FDE denotes the best ADE/FDE performance of trajectories with top 20 probabilities. We observe that the baseline Trajectron++ is the current SOTA method with the strong representation ability of social interactions. And with a better latent variable, our DisDis method achieves further improvement on the Rank $1$ with no cost of the performance of Rank $20$. It is because DisDis can select better predicted trajectories in higher probability. Furthermore, we fix the gradient computation problem in Trajectron++ as the new baseline, as Trajectron++*, and find that the improvement of DisDis is consistent. We also report the original ``best of 20'' performance in Table~\ref{table: bestof20}.

\subsection{Qualitative Evaluation} 

We qualitatively analyze how our DisDis method learns the personalized latent variable and improves motion predictions. Figure~\ref{fig: results_distribution} visualizes the learned latent distributions of our method and baseline method in 4 different scenarios including a pedestrian walking alone; pedestrians walking in parallel; pedestrians following the people ahead; and pedestrians meeting from different directions. We plot 20 top rank trajectories to represent the learned distributions. For all scenarios, we observe that our predicted trajectories are closer to ground-truth trajectories than the ones of baseline, which indicates more discriminative latent distributions. Taking the first image as the example, our predicted trajectories are obviously more concentrative than the ones of baseline method.  

In Figure~\ref{fig: tsne}, we compare the t-SNE visualization of latent variables. Points denote latent variables of trajectories and colors denote different pedestrians. Our method can learn more personalized latent variables than the baseline, e.g., the red and the black have their personalized motion patterns, but the baseline can't distinguish any pedestrian. Despite having learned the personalized latent variables, in (a), the last three share similar motion  patterns. It makes sense because people might have common habits.

Besides, we also provide the visualization examples with different scenes in the both ETH and UCY datasets in Figure~\ref{fig: results_examples}. We plot the most-likely predictions of our DisDis method and the baseline Trajectron++~\cite{salzmann2020trajectron++}. We observe that predicted trajectories are more similar to ground-truth trajectories, which demonstrates our method can obtain better prediction with higher probability. 
For the man in brown in the third scene of the ZARA2 dataset whose history trajectory shows a tendency to go straight, our method still learns the latent motion pattern turning left. And the crowded scene of UNIV demonstrates the effectiveness of our method to learn the discriminative distribution in the complex environment.

\section{Conclusion}

In this paper, we have proposed a distribution discrimination (DisDis) method to distinguish latent distribution via self-supervised contrastive learning. The DisDis method encourages the model to learn a discriminative latent variable space, where the trajectories with the same motion pattern have similar latent variable distribution while negative trajectories are pushed away. To evaluate the learned latent distribution, we further propose a probability cumulative minimum distance (PCMD) curve as the metric of stochastic trajectory prediction methods, which cumulatively calculates the minimum distance on the sorted probabilities. Finally, we show that our DisDis method could be integrated with existing stochastic predictors and obtain improvement in the learning of latent distribution.

{\small
\bibliographystyle{ieee_fullname}
\bibliography{trajectory}

\begin{thebibliography}{10}\itemsep=-1pt

\bibitem{alahi2016social}
Alexandre Alahi, Kratarth Goel, Vignesh Ramanathan, Alexandre Robicquet, Li
  Fei-Fei, and Silvio Savarese.
\newblock Social lstm: Human trajectory prediction in crowded spaces.
\newblock In {\em CVPR}, pages 961--971, 2016.

\bibitem{antonini2006discrete}
Gianluca Antonini, Michel Bierlaire, and Mats Weber.
\newblock Discrete choice models of pedestrian walking behavior.
\newblock {\em Transportation Research Part B: Methodological}, 40(8):667--687,
  2006.

\bibitem{chen2016variational}
Xi Chen, Diederik~P Kingma, Tim Salimans, Yan Duan, Prafulla Dhariwal, John
  Schulman, Ilya Sutskever, and Pieter Abbeel.
\newblock Variational lossy autoencoder.
\newblock {\em arXiv preprint arXiv:1611.02731}, 2016.

\bibitem{Fang_2020_CVPR}
Liangji Fang, Qinhong Jiang, Jianping Shi, and Bolei Zhou.
\newblock Tpnet: Trajectory proposal network for motion prediction.
\newblock In {\em CVPR}, June 2020.

\bibitem{felsen2018will}
Panna Felsen, Patrick Lucey, and Sujoy Ganguly.
\newblock Where will they go? predicting fine-grained adversarial multi-agent
  motion using conditional variational autoencoders.
\newblock In {\em ECCV}, pages 732--747, 2018.

\bibitem{fernando2018soft+}
Tharindu Fernando, Simon Denman, Sridha Sridharan, and Clinton Fookes.
\newblock Soft+ hardwired attention: An lstm framework for human trajectory
  prediction and abnormal event detection.
\newblock {\em Neural Networks}, 108:466--478, 2018.

\bibitem{goodfellow2014generative}
Ian Goodfellow, Jean Pouget-Abadie, Mehdi Mirza, Bing Xu, David Warde-Farley,
  Sherjil Ozair, Aaron Courville, and Yoshua Bengio.
\newblock Generative adversarial nets.
\newblock In {\em NeurIPS}, pages 2672--2680, 2014.

\bibitem{gupta2018social}
Agrim Gupta, Justin Johnson, Li Fei-Fei, Silvio Savarese, and Alexandre Alahi.
\newblock Social gan: Socially acceptable trajectories with generative
  adversarial networks.
\newblock In {\em CVPR}, pages 2255--2264, 2018.

\bibitem{He_2020_CVPR}
Kaiming He, Haoqi Fan, Yuxin Wu, Saining Xie, and Ross Girshick.
\newblock Momentum contrast for unsupervised visual representation learning.
\newblock In {\em CVPR}, June 2020.

\bibitem{helbing1995social}
Dirk Helbing and Peter Molnar.
\newblock Social force model for pedestrian dynamics.
\newblock {\em Physical Review E}, 51(5):4282, 1995.

\bibitem{huang2019stgat}
Yingfan Huang, HuiKun Bi, Zhaoxin Li, Tianlu Mao, and Zhaoqi Wang.
\newblock Stgat: Modeling spatial-temporal interactions for human trajectory
  prediction.
\newblock In {\em ICCV}, pages 6272--6281, 2019.

\bibitem{ivanovic2019trajectron}
Boris Ivanovic and Marco Pavone.
\newblock The trajectron: Probabilistic multi-agent trajectory modeling with
  dynamic spatiotemporal graphs.
\newblock In {\em ICCV}, pages 2375--2384, 2019.

\bibitem{kingma2014auto}
Diederik~P Kingma and Max Welling.
\newblock Auto-encoding variational bayes.
\newblock In {\em ICLR}, 2014.

\bibitem{kosaraju2019social}
Vineet Kosaraju, Amir Sadeghian, Roberto Mart{\'\i}n-Mart{\'\i}n, Ian Reid,
  Hamid Rezatofighi, and Silvio Savarese.
\newblock Social-bigat: Multimodal trajectory forecasting using bicycle-gan and
  graph attention networks.
\newblock In {\em NeurIPS}, pages 137--146, 2019.

\bibitem{lee2017desire}
Namhoon Lee, Wongun Choi, Paul Vernaza, Christopher~B Choy, Philip~HS Torr, and
  Manmohan Chandraker.
\newblock Desire: Distant future prediction in dynamic scenes with interacting
  agents.
\newblock In {\em CVPR}, pages 336--345, 2017.

\bibitem{lee2016predicting}
Namhoon Lee and Kris~M Kitani.
\newblock Predicting wide receiver trajectories in american football.
\newblock In {\em WACV}, pages 1--9, 2016.

\bibitem{lerner2007crowds}
Alon Lerner, Yiorgos Chrysanthou, and Dani Lischinski.
\newblock Crowds by example.
\newblock In {\em Computer Graphics Forum}, volume~26, pages 655--664, 2007.

\bibitem{liang2019peeking}
Junwei Liang, Lu Jiang, Juan~Carlos Niebles, Alexander~G Hauptmann, and Li
  Fei-Fei.
\newblock Peeking into the future: Predicting future person activities and
  locations in videos.
\newblock In {\em CVPR}, pages 5725--5734, 2019.

\bibitem{mangalam2020not}
Karttikeya Mangalam, Harshayu Girase, Shreyas Agarwal, Kuan-Hui Lee, Ehsan
  Adeli, Jitendra Malik, and Adrien Gaidon.
\newblock It is not the journey but the destination: Endpoint conditioned
  trajectory prediction.
\newblock In {\em ECCV}, 2020.

\bibitem{mohamed2020social}
Abduallah Mohamed, Kun Qian, Mohamed Elhoseiny, and Christian Claudel.
\newblock Social-stgcnn: A social spatio-temporal graph convolutional neural
  network for human trajectory prediction.
\newblock In {\em CVPR}, 2020.

\bibitem{nikhil2018convolutional}
Nishant Nikhil and Brendan Tran~Morris.
\newblock Convolutional neural network for trajectory prediction.
\newblock In {\em ECCVW}, pages 0--0, 2018.

\bibitem{oord2018representation}
Aaron van~den Oord, Yazhe Li, and Oriol Vinyals.
\newblock Representation learning with contrastive predictive coding.
\newblock {\em arXiv preprint arXiv:1807.03748}, 2018.

\bibitem{pellegrini2010improving}
Stefano Pellegrini, Andreas Ess, and Luc Van~Gool.
\newblock Improving data association by joint modeling of pedestrian
  trajectories and groupings.
\newblock In {\em ECCV}, pages 452--465, 2010.

\bibitem{robicquet2016learning}
Alexandre Robicquet, Amir Sadeghian, Alexandre Alahi, and Silvio Savarese.
\newblock Learning social etiquette: Human trajectory understanding in crowded
  scenes.
\newblock In {\em ECCV}, pages 549--565, 2016.

\bibitem{rodriguez2011data}
Mikel Rodriguez, Josef Sivic, Ivan Laptev, and Jean-Yves Audibert.
\newblock Data-driven crowd analysis in videos.
\newblock In {\em ICCV}, pages 1235--1242, 2011.

\bibitem{rudenko2020human}
Andrey Rudenko, Luigi Palmieri, Michael Herman, Kris~M Kitani, Dariu~M Gavrila,
  and Kai~O Arras.
\newblock Human motion trajectory prediction: A survey.
\newblock {\em IJRR}, 2020.

\bibitem{Sadeghian2019CVPR}
Amir Sadeghian, Vineet Kosaraju, Ali Sadeghian, Noriaki Hirose, Hamid
  Rezatofighi, and Silvio Savarese.
\newblock Sophie: An attentive gan for predicting paths compliant to social and
  physical constraints.
\newblock In {\em CVPR}, June 2019.

\bibitem{sadeghian2019sophie}
Amir Sadeghian, Vineet Kosaraju, Ali Sadeghian, Noriaki Hirose, Hamid
  Rezatofighi, and Silvio Savarese.
\newblock Sophie: An attentive gan for predicting paths compliant to social and
  physical constraints.
\newblock In {\em CVPR}, pages 1349--1358, 2019.

\bibitem{salzmann2020trajectron++}
Tim Salzmann, Boris Ivanovic, Punarjay Chakravarty, and Marco Pavone.
\newblock Trajectron++: Dynamically-feasible trajectory forecasting with
  heterogeneous data.
\newblock In {\em ECCV}, 2020.

\bibitem{sohn2015learning}
Kihyuk Sohn, Honglak Lee, and Xinchen Yan.
\newblock Learning structured output representation using deep conditional
  generative models.
\newblock In {\em NeurIPS}, pages 3483--3491, 2015.

\bibitem{Sun_2020_CVPR}
Hao Sun, Zhiqun Zhao, and Zhihai He.
\newblock Reciprocal learning networks for human trajectory prediction.
\newblock In {\em CVPR}, June 2020.

\bibitem{Sun_2020_CVPR2}
Jianhua Sun, Qinhong Jiang, and Cewu Lu.
\newblock Recursive social behavior graph for trajectory prediction.
\newblock In {\em CVPR}, June 2020.

\bibitem{tang2019multiple}
Charlie Tang and Russ~R Salakhutdinov.
\newblock Multiple futures prediction.
\newblock In {\em NeurIPS}, pages 15398--15408, 2019.

\bibitem{tay2008modelling}
Meng Keat~Christopher Tay and Christian Laugier.
\newblock Modelling smooth paths using gaussian processes.
\newblock In {\em Field and Service Robotics}, pages 381--390, 2008.

\bibitem{tian2019contrastive}
Yonglong Tian, Dilip Krishnan, and Phillip Isola.
\newblock Contrastive multiview coding.
\newblock {\em arXiv preprint arXiv:1906.05849}, 2019.

\bibitem{treuille2006continuum}
Adrien Treuille, Seth Cooper, and Zoran Popovi{\'c}.
\newblock Continuum crowds.
\newblock In {\em TOG}, volume~25, pages 1160--1168, 2006.

\bibitem{vemula2018social}
Anirudh Vemula, Katharina Muelling, and Jean Oh.
\newblock Social attention: Modeling attention in human crowds.
\newblock In {\em ICRA}, pages 1--7, 2018.

\bibitem{wang2007gaussian}
Jack~M Wang, David~J Fleet, and Aaron Hertzmann.
\newblock Gaussian process dynamical models for human motion.
\newblock {\em TPAMI}, 30(2):283--298, 2007.

\bibitem{wu2018unsupervised}
Zhirong Wu, Yuanjun Xiong, Stella~X Yu, and Dahua Lin.
\newblock Unsupervised feature learning via non-parametric instance
  discrimination.
\newblock In {\em CVPR}, pages 3733--3742, 2018.

\bibitem{xue2018ss}
Hao Xue, Du~Q Huynh, and Mark Reynolds.
\newblock Ss-lstm: A hierarchical lstm model for pedestrian trajectory
  prediction.
\newblock In {\em WACV}, pages 1186--1194, 2018.

\bibitem{yamaguchi2011you}
Kota Yamaguchi, Alexander~C Berg, Luis~E Ortiz, and Tamara~L Berg.
\newblock Who are you with and where are you going?
\newblock In {\em CVPR}, pages 1345--1352, 2011.

\bibitem{yu2020spatio}
Cunjun Yu, Xiao Ma, Jiawei Ren, Haiyu Zhao, and Shuai Yi.
\newblock Spatio-temporal graph transformer networks for pedestrian trajectory
  prediction.
\newblock In {\em ECCV}, 2020.

\bibitem{zhang2019sr}
Pu Zhang, Wanli Ouyang, Pengfei Zhang, Jianru Xue, and Nanning Zheng.
\newblock Sr-lstm: State refinement for lstm towards pedestrian trajectory
  prediction.
\newblock In {\em CVPR}, pages 12085--12094, 2019.

\bibitem{zhao2019infovae}
Shengjia Zhao, Jiaming Song, and Stefano Ermon.
\newblock Infovae: Balancing learning and inference in variational
  autoencoders.
\newblock In {\em AAAI}, pages 5885--5892, 2019.

\bibitem{zhao2019multi}
Tianyang Zhao, Yifei Xu, Mathew Monfort, Wongun Choi, Chris Baker, Yibiao Zhao,
  Yizhou Wang, and Ying~Nian Wu.
\newblock Multi-agent tensor fusion for contextual trajectory prediction.
\newblock In {\em CVPR}, pages 12126--12134, 2019.

\end{thebibliography}
}

\clearpage
\begin{appendix}
\section{Optimizing MI with DisDis.} 
In (6), we define the $ h(\mathbf{z},\mathbf{x}) $ with the ratio between $ p_{\theta}(\mathbf{z}|\mathbf{x})$ and $p_{\theta}(\mathbf{z})$. Thus, the objective function $\mathcal{L}_{3} $ becomes a lower bound of the mutual information as been proved in~\cite{oord2018representation}:
\begin{equation}
  \begin{aligned}
   \mathcal{L}_{3}(\theta) &= -\mathbb{E}_{\mathbf{z}} \log \bigg[ \frac{ \frac{p_{\theta}(\mathbf{z}|\mathbf{x})}{p_{\theta}(\mathbf{z})}} {  \frac{ p_{\theta}(\mathbf{z}|\mathbf{x})}{ p_{\theta}(\mathbf{z})}  + \sum_{\mathbf{z}_i \neq \mathbf{z}} \frac{p_{\theta}(\mathbf{z}_i|\mathbf{x})}{p_{\theta}(\mathbf{z}_i)}     }\bigg]   \\
      & = \mathbb{E}_{\mathbf{z}} \log \bigg[ 1+ \frac{p_{\theta}(\mathbf{z})}{p_{\theta}(\mathbf{z}|\mathbf{x})} \sum_{\mathbf{z}_i \neq \mathbf{z}} \frac{p_{\theta}(\mathbf{z}_i|\mathbf{x})}{p_{\theta}(\mathbf{z}_i)} \bigg]   \\
      &\approx  \mathbb{E}_{\mathbf{z}} \log \bigg[ 1+ \frac{p_{\theta}(\mathbf{z})}{p_{\theta}(\mathbf{z}|\mathbf{x})} (N'-1) \mathbb{E}_{\mathbf{z}_i} \frac{p_{\theta}(\mathbf{z}_i|\mathbf{x})}{p_{\theta}(\mathbf{z}_i)} \bigg] \\
      &= \mathbb{E}_{\mathbf{z}} \log \bigg[ 1+ \frac{p_{\theta}(\mathbf{z})}{p_{\theta}(\mathbf{z}|\mathbf{x})} (N'-1)  \bigg] \\
      &\geq \mathbb{E}_{\mathbf{z}} \log \bigg[ \frac{p_{\theta}(\mathbf{z})}{p_{\theta}(\mathbf{z}|\mathbf{x})} N'  \bigg] \\
      & = -\mathbb{E}_{\mathbf{z}} \log \frac{p_{\theta}(\mathbf{z}|\mathbf{x})}{p_{\theta}(\mathbf{z})} +\log(N') \\
      & = -I(\mathbf{z},\mathbf{x}) +\log(N').
  \end{aligned}
\end{equation}
Conventional self-supervised learning methods increase the mutual information between the trajectory and its variants under different views (e.g. rotation), such as CMC~\cite{tian2019contrastive} and Moco~\cite{He_2020_CVPR}. The loss function of self-supervised learning $\mathcal{L}_{self} $ can be written as:
\begin{equation}
  \begin{aligned}
   \mathcal{L}_{self}(\theta) = -\mathbb{E}_{\mathbf{x}'} \log \bigg[ \frac{ \frac{p_{\theta}(\mathbf{x}'|\mathbf{x})}{p_{\theta}(\mathbf{x}')}} {  \frac{ p_{\theta}(\mathbf{x}'|\mathbf{x})}{ p_{\theta}(\mathbf{x}')}  + \sum\limits_{\mathbf{x}'_i \in \mathbf{X}_{neg}  } \frac{p_{\theta}(\mathbf{x}'_i|\mathbf{x})}{p_{\theta}(\mathbf{x}'_i)}     }\bigg],
  \end{aligned}
\end{equation}
where $\mathbf{x}' $ denotes the variant under different views. 
The main difference between DisDis and these self supervised learning methods is that DisDis optimizes the mutual information between trajectory and the latent distribution, which encourages the trajectory embeddings to be close to the ones of with the same latent distribution, instead of its variants under different views.
\begin{figure*}[!t]
\centering
\includegraphics[width=0.95\textwidth]{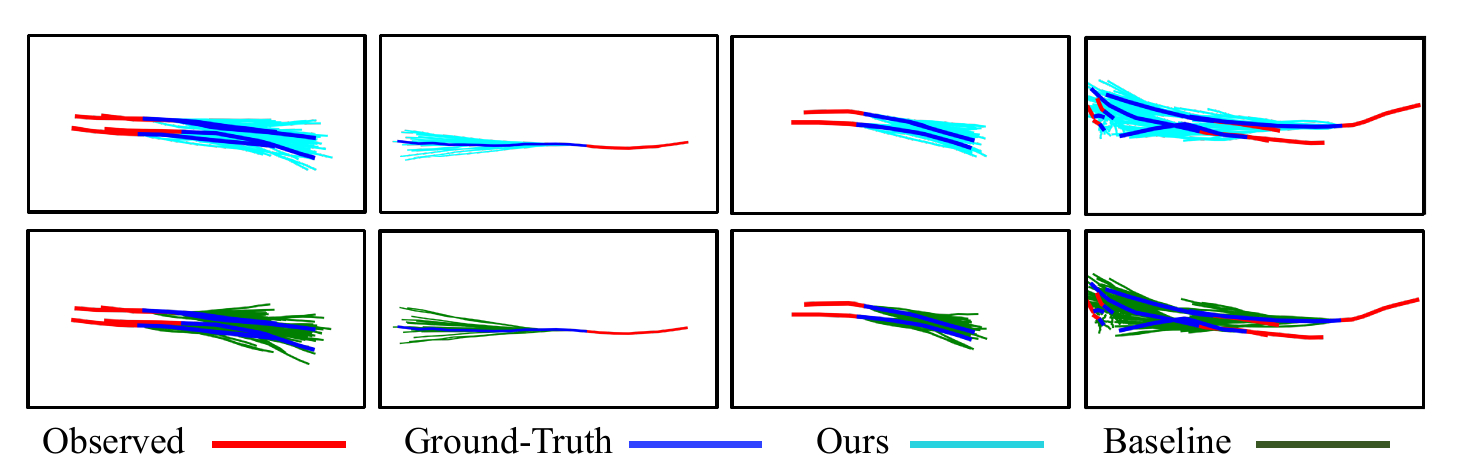}
    \caption{Illustration of the learned distributions. We additionally provide some comparisons of learned distributions between our method and Trajectron++~\cite{salzmann2020trajectron++} baseline. (Best viewed in color.)}
\label{fig: results_distribution}
\vspace{-0.3cm}
\end{figure*}

\begin{figure*}[!t]
\centering
\includegraphics[width=0.96\textwidth]{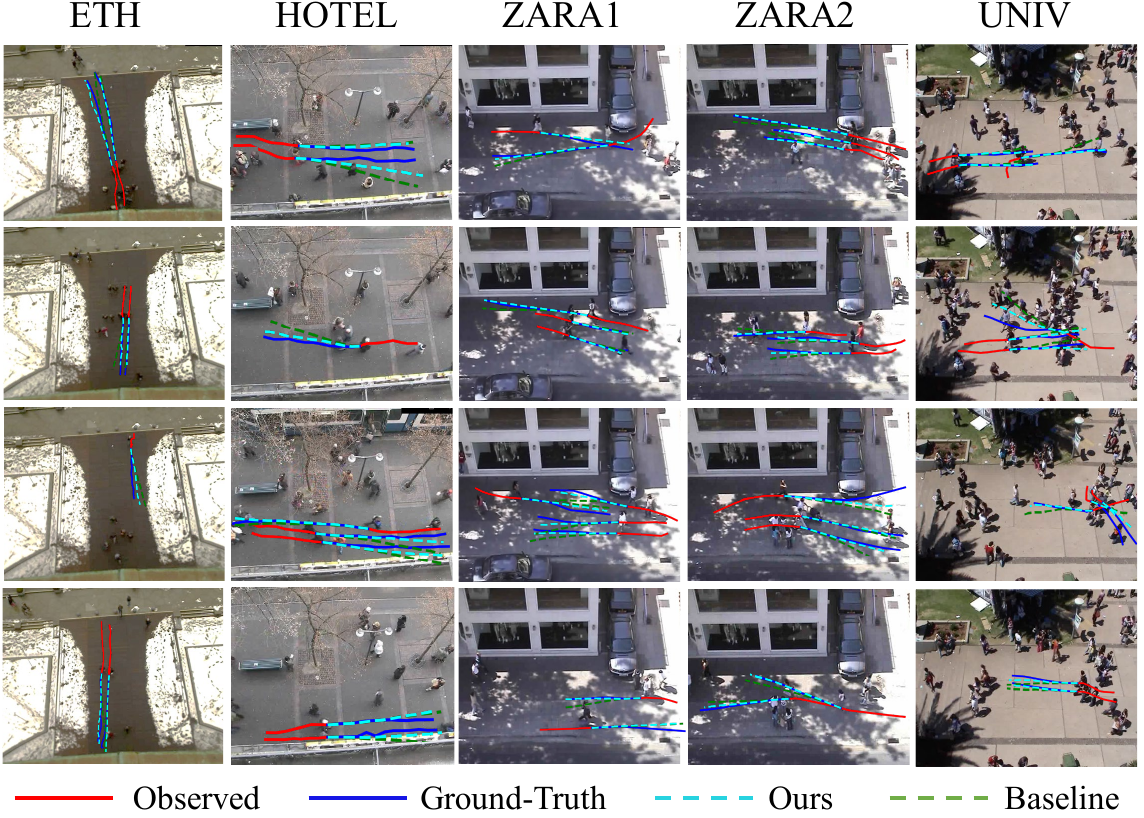}
   \caption{ Visualization examples of our DisDis method and baseline Trajectron++~\cite{salzmann2020trajectron++} method in the different scenes. We provide some additional visualization examples for all five scenes in the both ETH and UCY datasets. (Best viewed in color.)}
   \label{fig: results_examples}
\vspace{-0.2cm}
\end{figure*}

\section{Detailed Derivations of PCMD.} 

Motivated by CMC curve in recognition tasks, the Y-axis of PCMD refers to the accuracy of the trajectory (the distance between prediction and ground-truth) while the X-axis refers to the distribution space of trajectory. We select trajectories according to the probability of $p(z \vert x)$, which means that we sort variables with different values according to probability. 

First, we need to sort the distribution of  $p(z \vert x)$ with the Lebesgue measure: Formally, given the distribution $p_\theta(\mathbf{z} \vert \mathbf{x}), \mathbf{z}\in\Omega$, we define a projection as
\begin{equation}
F(\tau) = \mathbb{E}_{\mathbf{z}\in \Omega} \mathbb{I}(p_\theta(\mathbf{z}\vert \mathbf{x}) \geq \tau),
\end{equation}
where the input denotes a selected probability value $\tau \in (\min p_\theta(\mathbf{z}\vert \mathbf{x}), \max p_\theta(\mathbf{z}\vert \mathbf{x}) ) $, $ \mathbb{I} $ is an indicator function, while the output $F(\tau) \in(0,\Vert\Omega \Vert) $ denotes the interval length of $\mathbf{z}$ satisfying conditions. $F(\tau)$ refers to the expectation of area where the probability function value is greater than the given probability $\tau$. Then the X-axis of PCMD can be calculated by $F(\tau)$. We define $k= \frac{F(\tau)}{\Vert\Omega \Vert}$ to normalize it into $(0,1) $, if latent space $\Omega $ is not finite.
We obtain the values of PCMD curve as:
\begin{equation}
PCMD(k) =\mathbb{E}_{\mathbf{x}} \min\{ \mathcal{D}(\mathbf{z}) \vert p_\theta(\mathbf{z}\vert \mathbf{x}) \geq \tau,\mathbf{z} \in \Omega\},
\end{equation}
where $ \mathcal{D}(\mathbf{z})$ denotes the ADE or FDE distance between the ground-truth and the sampled predicted trajectories based on $\mathbf{z}$. If the latent space $\Omega $ is finite, we can definde $\Omega^{*} $ as follows:
\begin{equation}
\mathbb{E}_{z \sim p(z) } p(z\in \Omega^{*}) \geq 1-\epsilon, \Vert \Omega^{*} \Vert \leq N .
\end{equation}
Then, we can re-organize $PCMD(k)$ among $\Omega^{*}$. 

However to calculate the continuous distribution is too difficult and nonnecessary, we can approximate $PCMD(k)$ with  Monte Carlo methods: we sample M variables $ Z= \{\mathbf{z}_i \in \Omega \vert i=1,2,\cdots,M \} $ and sort these variables with the probability $p_\theta(\mathbf{z}_i\vert \mathbf{x})$ from large to small to obtain $Z_{sort} = \{\mathbf{z}^*_i \} $. Then, the values of PCMD are calculated as:
\begin{equation}
PCMD(k) =\mathbb{E}_{\mathbf{x}} \min\{\mathcal{D}(\mathbf{z}) \vert z \in \{ z^*_1,z^*_2, \cdots, z^*_m\}\},
\end{equation}
where $k = \frac{m}{M}$ denotes the ranking rate. More importantly, the condition $\Vert\Omega \Vert \to \infty $ can be ignored when we discretely approximate it by the Monte Carlo methods. 

\section{Ablation Study for Loss Function.}
We conduct an ablation study for loss functions on the UNIV dataset, which shows the effectiveness of our method. $\mathcal{L}_{1}$ and $\mathcal{L}_{2}$ make a whole of CVAE, so the model breaks down and performs worst when using the loss function without $\mathcal{L}_{2}$.
\renewcommand\tabcolsep{5pt}
\begin{table}[h]
\begin{center}
\caption{Ablation studies of losses on the UNIV dataset. }
\vspace{1pt}
\begin{tabular}{c|cccc}
\hline
M=80	& DisDis & w/o $\mathcal{L}_{2}$ & w/o $\mathcal{L}_{3}$  \\
\hline
ADE   & 0.36 & 0.82 & 0.38 \\ 
FDE   & 0.92 & 2.07 & 0.96 \\
\hline
\end{tabular}
\label{tab: univ}
\end{center}
\end{table}
\vspace{-20pt}

\section{More Visual Results.} 

To have an intuitive understanding of our DisDis method, we qualitatively compare the the visualizations of our method and baseline Trajectron++~\cite{salzmann2020trajectron++}. The more visual results are presented in Figure~\ref{fig: results_distribution} and Figure~\ref{fig: results_examples}. We see our method can obtain better prediction with higher probability. It indicates DisDis can work better when we only select a few predictions.

\end{appendix}

\end{document}